\documentclass[lettersize,journal]{IEEEtran}
\usepackage{amsmath,amssymb,amsfonts}
\usepackage{algorithmic}
\usepackage{algorithm2e}
\usepackage{array}
\usepackage[caption=false,font=normalsize,labelfont=sf,textfont=sf]{subfig}
\usepackage{textcomp}
\usepackage{stfloats}
\usepackage{url}
\usepackage{verbatim}
\usepackage{graphicx}
\usepackage{bbm}

\def\BibTeX{{\rm B\kern-.05em{\sc i\kern-.025em b}\kern-.08em
    T\kern-.1667em\lower.7ex\hbox{E}\kern-.125emX}}
\usepackage{balance}
\begin{document}
\title{Two-phase Dual COPOD Method for Anomaly Detection in Industrial Control System}
\author{Emmanuel Aboah Boateng,  \textit{Student Member, IEEE}, and J.W. Bruce, \textit{Senior Member, IEEE}
\thanks{

The authors are with the Department of Electrical and Computer Engineering, Tennessee Technological University, Cookeville, TN, USA.
}
}


\maketitle

\begin{abstract}
Critical infrastructures like water treatment facilities and power plants depend on industrial control systems (ICS) for monitoring and control, making them vulnerable to cyber attacks and system malfunctions. Traditional ICS anomaly detection methods lack transparency and interpretability, which make it difficult for practitioners to understand and trust the results. This paper proposes a two-phase dual Copula-based Outlier Detection (COPOD) method that addresses these challenges. The first phase removes unwanted outliers using an empirical cumulative distribution algorithm, and the second phase develops two parallel COPOD models based on the output data of phase 1. The method is based on empirical distribution functions, parameter-free, and provides interpretability by quantifying each feature's contribution to an anomaly. The method is also computationally and memory-efficient, suitable for low- and high-dimensional datasets. Experimental results demonstrate superior performance in terms of F1-score and recall on three open-source ICS datasets, enabling real-time ICS anomaly detection.
\end{abstract}


\section{Introduction}
\label{sec:introduction}
\IEEEPARstart{M}{odern} critical infrastructures like water treatment facilities, oil refineries, power grids, and nuclear and thermal power plants all include industrial control systems (ICS) and are used to control and monitor a physical process. A system known as an ICS is created by combining computational, and communication components, sensors, actuators, Programmable Logic Controllers (PLCs), Human Machine Interfaces (HMIs), and Supervisory Control and Data Acquisition (SCADA) systems are some of the devices and subsystems that make up an ICS. 

\begin{figure}[h]
\centerline{\includegraphics[width=\columnwidth]{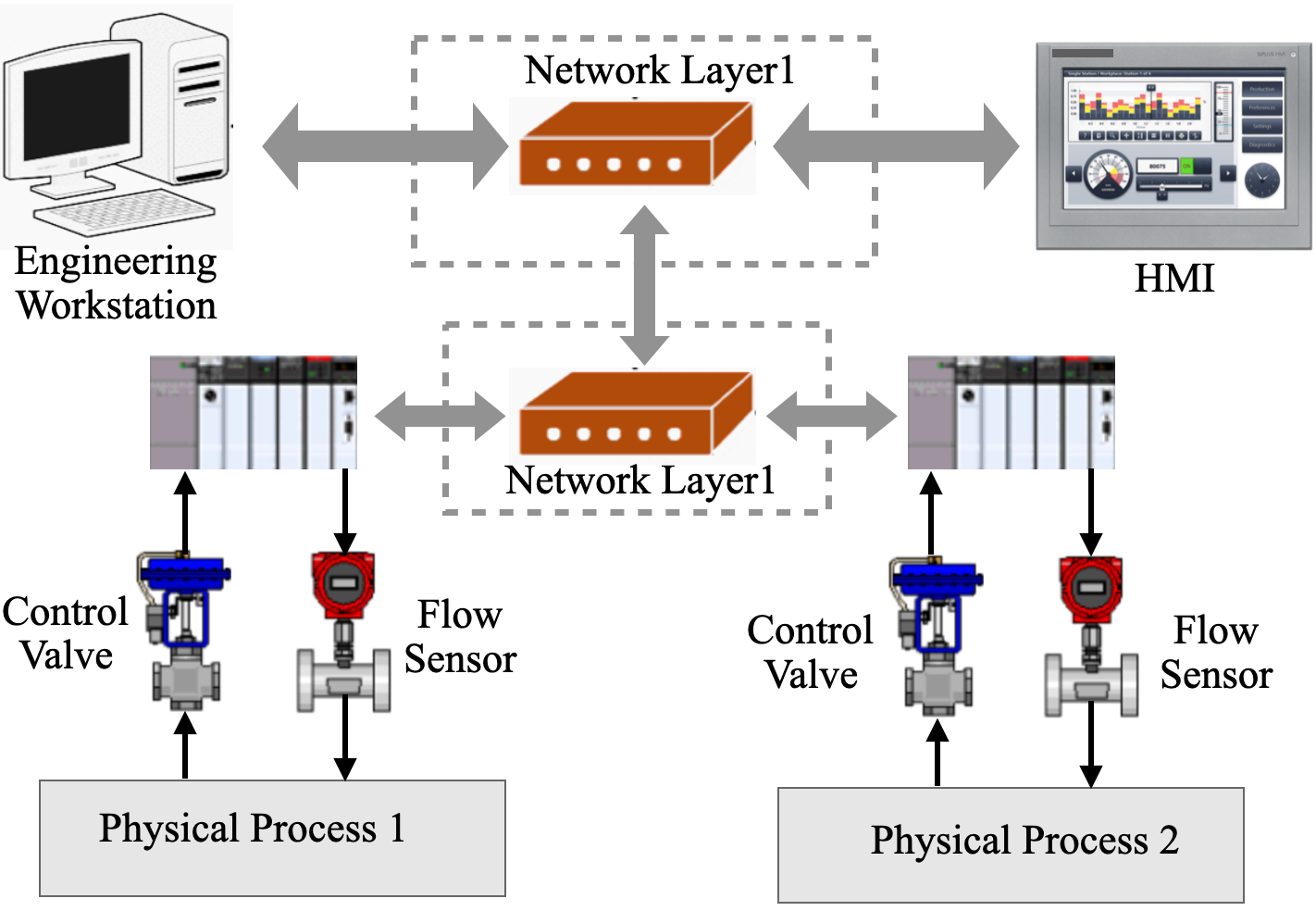}}
\caption{Conceptual View of Industrial Control Systems (ICS).}
\label{Figure1}
\end{figure}

For a given ICS setup, the physical layer's field devices, or sensors and actuators, control and manage the underlying industrial process. The distributed control layer's PLCs get information about the process's present condition via sensors. Pumps, valves, generators, and circuit breakers are examples of actuators that receive control actions from the PLCs and carry them out. For the purpose of executing human-assisted control actions, other devices, such as the SCADA and HMIs at the supervisory control layer, enable communication between a plant operator and the PLCs \cite{mr2021machine}. Figure \ref{Figure1} shows a conceptual view of a typical ICS.

Because of the deployment of ICSs in critical infrastructure, ICSs are desirable targets for adversaries. Any of the ICS subsystems may be compromised by a skilled adversary who would then be able to manipulate the actuators or sensor readings over time until a nefarious objective is achieved. ICSs may experience service interruptions or material losses as a result of malicious cyberattacks, which could have a detrimental impact on the quality of life in societies. A significant instance is the 2010 Stuxnet-based cyberattack against the Iranian nuclear program \cite{karnouskos2011stuxnet}. The worm used pertinent data from the ICS to harm the centrifuges inside the factory by repeatedly altering their rotation speed after infecting the Programmable Logic Controllers (PLCs). Another such is the cyberattack in late 2015 \cite{alert2016cyber} that targeted the electrical grid in Ukraine. Three local power distribution businesses had their information systems compromised by the attackers using the BlackEnergy malware. 

Machine learning-based intrusion detection solutions for ICSs have been developed in response to the rising frequency and complexity of these attacks \cite{aboah2022plc, aboahboateng2021anomaly, liu2018novel}. A common and effective method among the several strategies suggested in the scientific literature is the one-class classification \cite{aboah2022plc, cao2016one}. In the training stage, one-class classification-based solutions create a model that embodies the ICS's typical behavior. The detection system employs that model during the production stage to check whether the behavior of the live system adheres to a required standard. Anomalies are deviations from the norm that are typically identified using classification error criteria. These algorithms may be particularly sensitive to unusual behavior, including zero-day anomalies or cyberattacks and malfunctioning hardware or sensors that might avoid detection.

Another key challenge of deploying machine learning techniques for ICS is the computational demands of the machine learning detection model \cite{vavra2021adaptive}. ICS technology has been around for several decades; as a result, deploying complex machine learning models with high computational requirements can be challenging. However, existing approaches are expensive to compute. Existing approaches performances are limited to low-dimensional data due to the curse of dimensionality. The majority of existing methods require the selection and tuning of hyperparameters. Sometimes the selection of the hyperparameters is arbitrary and subjective, producing various results during different experiments of the same ICS setup. 

Model interpretability is a crucial challenge in ICS anomaly detection. The majority of the existing literature makes use of complex machine learning techniques \cite{elnour2020dual, aboah2022plc, aboahboateng2021anomaly} or an ensemble of different anomaly detection algorithms \cite{zhao2015ensemble, ganaie2021ensemble, aboah2023unsupervised} in an effort achieve high anomaly detection performance. However, these complex anomaly detection techniques suffer from model interpretability. In ICS setup, it is essential to understand the root cause of anomalies to make informed decisions during ICS troubleshooting and during ICS forensic analysis. Because ICS control critical infrastructure in our society, anomaly detection models must be capable of providing reasons for flagging any system operation as an anomaly.

False alarms are also a challenge in ICS anomaly detection. To preserve system availability and operational continuity, every anomaly detection system must lower the erroneous categorization of ICS routine procedures. A false alarm could occur due to the selection of an improper machine learning approach for ICS anomaly detection. For example, supervised learning has been used by the vast majority of researchers for ICS anomaly detection \cite{kravchik2018detecting, liu2018novel}. Because this supervised anomaly detection method is unable to detect unidentified cyberattacks, huge, up-to-date databases, including every possible cyberattack, are required to train the classification model. Such requirements are infeasible in practice. Unsupervised machine learning techniques have also been used in ICS anomaly detection \cite{aboah2022plc, elnour2020dual} in an effort to account for unknown anomalies. Unsupervised anomaly detection algorithms assume that all training data are a true representation of the normal operation of the ICS because anomalies are unknown ahead of time. The assumption may not hold for complex ICS with several sensors and actuators. Sensor and actuator readings may be noisy. If noisy data is used in training anomaly detection models, their predictions may not be a
true reflection of the normal ICS operation. There is a need to remove the noise and any apparent outliers in the data before training.

This work tackles the aforementioned challenges by proposing a novel unsupervised Two-phase Dual (TPD) Copula-Based Outlier Detection (COPOD) method for ICS anomaly detection. The proposed TPD COPOD is an anomaly detection method that consists of two sequential modeling stages and a dual (parallel) modeling stage. The first phase of the method makes use of Empirical Cumulative Distribution Functions (ECOD) to remove any obvious noise in the training dataset in order to substantiate the hypothesis that the training dataset truly represents the normal operation of a given ICS dataset. ECOD uses information about the data distribution of the training dataset to identify anomalies based on the assumption that anomalies are mostly isolated events located in the tails of a distribution \cite{li2022ecod}. The second phase consists of a dual COPOD architecture that utilizes the normal process data (output) of phase 1 to develop two COPOD models. The raw normal process data is used to develop the first COPOD model. The second COPOD model is developed using normalized process data (excluding discrete variables). COPOD algorithm uses the normal dataset to formulate an empirical copula, and then uses the copula to predict the tail probabilities of each observation in the dataset \cite{li2020copod}. The magnitude of the tail probabilities represents the degree to which an observation is an outlier. A decision function strategy is also introduced for assigning final anomaly scores for the second phase. The motivation for the proposed architecture is to be able to exploit ICS data by examining the data using two latent representations to extract useful information to minimize overfitting and improve the model’s anomaly detection capabilities.

The contributions of this work can be summarized below:
\begin{enumerate}
\item	Propose first known two-phase dual COPOD method for ICS anomaly detection;
\item   Introduce an efficient and scalable anomaly detection method suitable for both low and high dimensional ICS datasets; and capable of real-time ICS anomaly detection; and
\item	Introduce a robust, parameter-free ICS anomaly detection method based on empirical distribution functions. The deterministic nature of all stages of the method leads to the mitigation of the challenges involved in hyperparameter selection.
\item   Propose a highly interpretable anomaly detection method that quantifies each feature's contribution towards an ICS anomaly
\end{enumerate}
The rest of this article is organized as follows. Section II presents an overview of the related works, followed by Sections III, which discusses the details of the experimental setup and various datasets used for developing the proposed anomaly detection method. Section IV presents the proposed method, and after that, section V presents the results and discussions. Finally, Section VI presents conclusions and provides recommendations for future work.

\section{Related Work}
Anomaly and intrusion detection in ICSs has been the subject of extensive research. System state prediction is the first stage in physics-based detection, as mentioned in \cite{giraldo2018survey}. For instance, some research employed linear dynamical system modeling \cite{mishra2016secure, murguia2016characterization} or autoregressive models \cite{ mashima2012evaluating}. The linear dynamical system proposed in \cite{mishra2016secure} is based on a secure state estimation algorithm that involves Kalman filters for attack detection and state estimation against sensor attacks in a noiseless dynamical system. The work in \cite{mashima2012evaluating} makes use of Autoregressive Moving Average models for meter management data anomaly detection. The ARMA model in \cite{mashima2012evaluating} is a linear process that assumes that the meter management data is stationary and when the data fluctuates, it does so uniformly around a particular time.  Unfortunately, the linearity of the modeled system, which is generally not satisfied in ICSs, is one of the assumptions made by both techniques.

Invariant-based techniques \cite{rahman2016multi, roth2016physical} and specification-based system modeling \cite{mishra2019modeling} have been shown to be efficient anomaly detection approaches. Specificity or the requirement that the solution is suited to the system and its operating circumstances is one of the key downsides of invariant-based techniques and specification-based systems.

Approaches based on statistical models in the context of anomaly detection typically start by fitting probability distributions to data points. Using the fitted models, the statistical models decide which points are outliers. Parametric and nonparametric methods are two main classes into which these techniques are typically divided. The main distinction is that parametric methods take the data to come from a parametric distribution under the assumption; hence fitting such a distribution entails learning the parameters of the assumed parametric distribution. Both linear regression and Gaussian mixture models (GMM) are used parametric techniques for anomaly detection \cite{yang2009outlier, satman2013new}. In contrast, nonparametric techniques do not rely on a parametric model of the data. Examples include histogram-based techniques (HBOS) \cite{goldstein2012histogram}, Kernel Density Estimation (KDE) \cite{pavlidou2014kernel}. After the model fitting process, parametric models for anomaly detection are typically quick to utilize; however, nonparametric models for anomaly detection can be more expensive to deal with. 

To obtain reliable anomaly detection results, ensemble-based methods integrate the output from different base outlier detectors. Notable works of ensemble based anomaly detection methods include feature bagging \cite{lazarevic2005feature}, isolation forest \cite{liu2008isolation}, locally selective methods \cite{zhao2019lscp}, and scalable unsupervised outlier detection \cite{zhao2021suod}. Feature bagging method employs a variety of sub-feature spaces. Isolation forests method combines data from multiple base trees. Locally selective combination in parallel outlier ensembles method dynamically selects the best base estimator for each data point. Scalable unsupervised outlier detection method employs numerous heterogeneous estimators. In general, ensemble-based approaches for anomaly detection frequently perform well in practice, even for high-dimensional datasets \cite{sun2016detecting, parveen2013evolving}. However, ensemble-based methods can require non-trivial tweaking, such as choosing the appropriate meta-detectors \cite{zhao2021automatic}. Ensemble anomaly detection methods are frequently difficult to interpret \cite{zhou2018ensemble}.

The COPOD anomaly detection algorithm is modeled after copulas for multivariate data distribution \cite{li2020copod}. Copulas are mathematical functions that let COPOD distinguish between the dependency structure of a given multivariate distribution and the marginal distributions of a dataset. Empirical Cumulative Distribution Functions (ECDF) are first calculated by the authors using a specified dataset. The empirical copula functions are then created using the (ECDF). Finally, the COPOD anomaly detection algorithm uses the empirical copula to approximate the tail probabilities and quantifies the probabilities as the anomaly scores of the data records. The authors in \cite{li2020copod} evaluated their proposed algorithm on 30 public outlier detection benchmark datasets. In each instance, 60\% of the data was used for training, and the remaining 40\% of the data was used for testing. Through extensive experiments conducted, the authors claim the detection algorithm is a state-of-the-art outlier detection algorithm in terms of detection accuracy and computational cost \cite{li2020copod}.

An unsupervised outlier detection algorithm known as ECOD, which is inspired by the fact that outliers are often the rare events that occur at the tails of a distribution, is proposed in \cite{li2022ecod}. The authors' method models each dimension in the dataset using a nonparametric statistical method. The proposed method calculates the ECDF per dimension for each data point in order to first estimate the underlying distribution of the input dataset. The algorithm then estimates the dataset's tail probabilities across all dimensions using the obtained ECDF. Finally, quantification of the tail probability is utilized to score the data records for outliers in the dataset. The authors conducted in-depth tests using 30 benchmark datasets and claimed that the ECOD algorithm outperforms 11 state-of-the-art baselines in terms of detection accuracy, model efficiency, and scalability. Despite their model's robust performance, the work in \cite{li2022ecod} made a simplistic unimodal assumption about the input dataset, and so the approach is not appropriate for multimodal distributions in which an outlier could be in neither left nor right tails.


The detection of anomalies in PLCs has also been done using NN-based prediction algorithms \cite{xiao2017nipad}. For the purpose of identifying PLC anomalies employing long short-term memory (LSTM), the authors in \cite{xiao2017nipad} presented a non-intrusive power-based anomaly detection approach. 
With the proposed models, accuracy of up to 98\% was obtained. However, identifying malicious code by monitoring the power consumption of the PLC is insufficient as a faulty power supply and a power electronics breakdown, can result in false positive readings \cite{sokolov2019applying}.

\section{Experiment Setup}
This section provides the details of the threat model for this study and the various experimental setups and datasets considered for developing and evaluating the proposed anomaly detection method.

\subsection{Threat Model}
In this work, it is assumed that an attacker is capable of physically compromising sensors and actuators inside the ICS, gaining remote access to the SCADA workstation, and gaining access to the networked control system. This work further assumes that the attacker is familiar with the targeted ICS, including the physical characteristics measured by each sensor and the effects of actuation commands. The objective of the attacker is to harm or alter the ICS activities using the aforementioned capabilities and prior information. This includes; manipulating actuator states through an attack akin to Man-In-the-Middle (MITM) attack in which the attacker sends orders to the actuator rather than a PLC; delivering forged sensor values to the PLC to influence the PLC to make poor judgments; and altering the PLC firmware with the intention to disable the ICS or alter the programmed logic.

\subsection{Datasets Description}
The proposed method is trained, evaluated, and compared with previous work using three open source ICS datasets, namely Secure Water Treatment (SWaT) \cite{mathur2016swat}, Water Distribution (WADI) \cite{adepu2016generalized}, and the Traffic Light (TLIGHT) datasets \cite{aboah2022plc, siemens1996s7}. This section provides a summary of the main properties of the datasets utilized in this work.

The SWaT testbed is a typical representation of the dynamic nature of Cyber-physical Systems (CPSs) used in our societies \cite{goh2017anomaly}. 
The dataset is subdivided into a 7-day fraction of regular operations, which serve as the training set, and a 4-day chunk of anomalous operations and 36 attacks produced using the attack model in \cite{adepu2016generalized}. 
The WADI dataset represents a scaled-down version of a large water distribution network in
a city. 
The data records in the WADI dataset, each with 123 attributes broken down into 69 sensor readings and 54 actuator states, were gathered over the course of 16 days. For the first 14 days, normal operating circumstances were recorded and divided into training (95\%) and validation (5\%) sets. The test set consists of the final two days containing 15 attacks interspersed with normal operating conditions. A total of 784,568 data records were used for training the proposed model and 172,800 data records were used for testing the proposed model.
The experimental setup for emulating the behavior of the traffic light (TLIGHT) system for the purposes of data collection is described in \cite{aboah2022plc, siemens1996s7}. 
The TLIGHT dataset consisting of normal and abnormal data records, is recorded for a period of four days. Overall, five test sets are presented in the TLIGHT dataset, with each set containing normal and anomalous TLIGHT system operations.

\section{Proposed Method}
This section describes the fundamental background of the two-phase dual COPOD anomaly detection technique. 

\subsection{Two-phase Dual COPOD Method}
\begin{figure}[h]
\centerline{\includegraphics[width=\columnwidth]{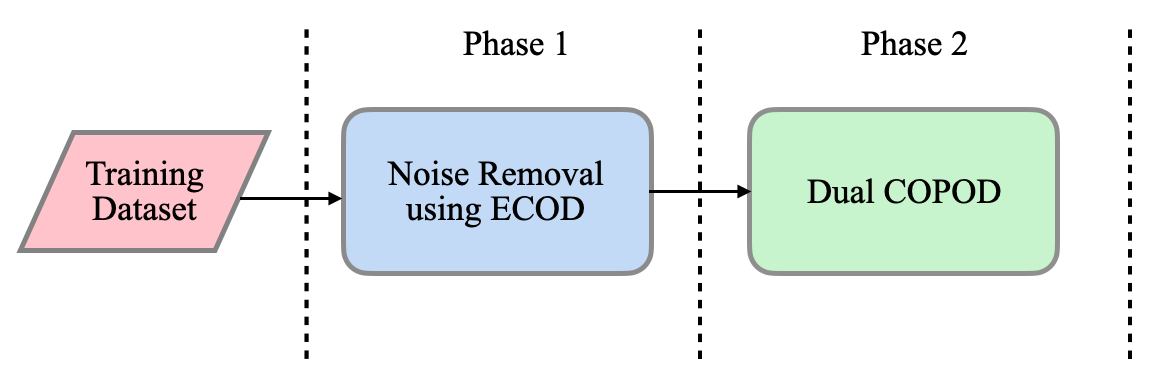}}
\caption{Two-phase Dual COPOD Method.}
\label{copod_fig2}
\end{figure}

Anomaly detection algorithms assume that all training data are true representations of the normal operation of a system because the anomalies are unknown ahead of time. The assumption may not hold for complex ICS with several sensors and actuators. Sensor and actuator readings may be noisy. If noisy data is used in training anomaly detection models, anomaly predictions may not be accurate. Therefore, noise and apparent outliers need to be removed in the data before training. To this end, a two-phase dual (TPD) COPOD method is proposed. The proposed TPD COPOD is an anomaly detection method that consists of two sequential modeling stages and a dual (parallel) modeling stage. Figure \ref{copod_fig2} shows the high-level overview of the TPD COPOD. The first phase is responsible for data cleaning and noise removal using ECOD. The second stage of the method is trained using the cleaned data from phase 1. The details of the individual stages of the method are discussed in the subsequent sections. 

\subsection{Phase 1: ECOD Model}
\begin{figure}[h]
\centerline{\includegraphics[width=9cm]{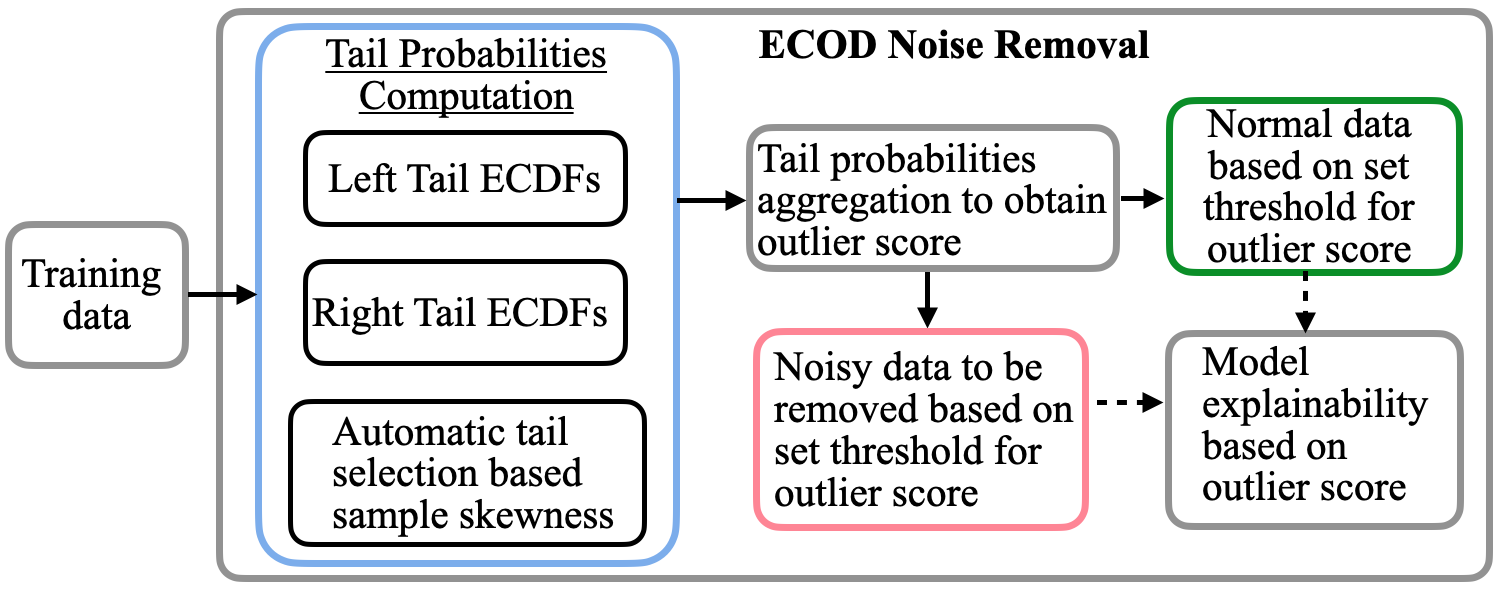}}
\caption{Phase 1 Architecture of the Proposed Method.}
\label{Figure4}
\end{figure}

Phase 1 of the method uses the ECOD algorithm \cite{li2022ecod} to remove anomalies and obvious noise in the training dataset so that the assumption about the training dataset being an authentic representation of the normal ICS operations holds. Anomalies are rare data points that occur in low-density areas of a probability distribution \cite{lazarevic2005feature, pokrajac2007incremental}. If the distribution is unimodal, these rare events are found in the distribution's tails. Determining the likelihood of finding a data point at least as "extreme" as $X_i$ in terms of tail probabilities forms the basis of the approach ECOD algorithm of phase 1. In this work, it is assumed that for a given ICS dataset, $n$ data points $X_1, X_2,...,X_n \in R^d$ are sampled independently and identically distributed. Figure \ref{Figure4} shows the architecture of phase 1 in the TPD COPOD method.

The joint cumulative distribution function over all $d$ dimensions is represented by $F:R^d \to [0, 1]$. Then, for a vector $z \in R^d$, let $z^{(j)}$ represent the $j-th$ entry. Hence by definition of a joint CDF, for any $x \in R^d$,
\begin{equation}
\label{eq1}
   F(x) = P(X^{(1)} \leq x^{(1)}, X^{(2)} \leq x^{(2)},..., X^{(d)} \leq x^{(d)})
\end{equation}

Regarding the left tail, \eqref{eq1} determines how extreme $X_i$  is. The smaller $F(X_i)$ is, the less likely a point $X$ sampled from the same distribution as $X_i$ will satisfy the inequality $X \leq X_i$. Similarly, $1-F(X_i)$ measures the extremeness of $X_i$ by focusing on the right tails of each dimension rather than the left tails. As a result, if either $F_{X}(X_i)$ or $1-F(X_i)$  is extremely small, then this suggests that $X_i$ corresponds to a rare event and is, therefore, likely to be an anomaly. To simplify, assume that a dataset's various dimensions are independent so that the combined CDF can be factored
\begin{equation*}
  F(x) = \prod_{j = 1}^{d} F^{(J)}(x^{(j)}) \quad \text{for} \quad x \in R^d,
\end{equation*}
where $F^{(j)} : R \to [0, 1]$ represents the univariate CDF of the j-th dimension. Now it is sufficient to notice that univariate CDFs can be accurately estimated by utilizing the ECDF. The left tail ECDF is
\begin{equation}
\label{eq2}
   \Hat{F}_{left}^{(j)}(z) = \frac{1}{n} \sum_{i = 1}^{n} \mathbbm{1}\{X_{i}^{(j)} \leq z\} \quad \text{for} \quad z \in R
\end{equation}
where $\mathbbm{1}\{\cdot\}$ is the indicator function that is 1 when its argument is true and is 0 otherwise. Similarly, the right tail ECDF is 
\begin{equation}
\label{eq3}
   \Hat{F}_{right}^{(j)}(z) = \frac{1}{n} \sum_{i = 1}^{n} \mathbbm{1}\{X_{i}^{(j)} \geq z\} \quad \text{for} \quad z \in R.
\end{equation}
For every point $X_{i}^{(j)}$, the tail probabilities $\Hat{F}_{left}^{(j)}(X_i^{(j)})$ and $\Hat{F}_{right}^{(j)}(X_i^{(j)})$ are aggregated by multiplying them together to achieve the final anomaly score $O_i \in [0, \infty)$. Higher $O_i$ means more likely to be an anomaly. 

The skewness of a dimension's distribution is used in aggregating the tail probabilities to determine whether the left or right tail probability should be used for a given dimension. The sample skewness coefficient, $\Tilde{\mu_3}$ of dimension $j$, for a given dataset can be derived as \cite{groeneveld1984measuring}
\begin{equation}
\label{eq4}
   \Tilde{\mu_3} = \frac{\sum_{i = 1}^{n}(X_i^{(j)} - \Bar{X^{(j)}})^{3}}{(n-1) \times \sigma^{3}}
\end{equation}
where $\sigma$, the standard deviation is
\begin{equation*}
\label{eq3}
  \sigma = \sqrt{\frac{\sum_{i = 1}^{n}(X_i^{(j)} - \Bar{X}^{(j)})^{2}}{(n-1)}}
\end{equation*}
and $\Bar{X}^{(j)}$ is the mean of dimension $j$'s distribution, and $\Bar{X}^{(j)} = \sum_{i = 1}^{n}X_i^{(j)}/n$.
When $\Tilde{\mu_3}< 0$, points on the left tail can be considered outliers, whereas when $\Tilde{\mu_3} > 0$, points on the right tail can be considered outliers.

The tail probabilities are transformed to log negative probabilities to calculate the final anomaly score per data point for a given training dataset. The log negative probability to the left and right tails are
\begin{equation}
\label{eq5}
  O_{left}(X_i) = -\sum_{j = 1}^{d} \log(\Hat{F}_{left}^{(j)}(X_i^{(j)})),
\end{equation}
and
\begin{equation}
\label{eq6}
  O_{right}(X_i) = -\sum_{j = 1}^{d} \log(\Hat{F}_{right}^{(j)}(X_i^{(j)})),
\end{equation}
respectively.The automated form of selecting the left or right tail of the $j-th$ dimension based on whether $\Tilde{\mu_3} < 0$  or $\Tilde{\mu_3} > 0$  is
\begin{equation}
\label{eq7}
\begin{split}
  O_{auto}(X_i) &= \sum_{j = 1}^{d}[ \mathbbm{1} \{\Tilde{\mu_3}  <0\} \log(\Hat{F}_{left}^{(j)}(X_i^{(j)})) \\
  &+  \mathbbm{1} \{\Tilde{\mu_3} \geq 0\} \log(\Hat{F}_{right}^{(j)}(X_i^{(j)}))].
  \end{split}
\end{equation}

The final anomaly score is calculated in the space of negative log probabilities, where a lower probability translates into a larger negative log probability and, as a result, a higher likelihood of being an anomaly. The largest magnitude of the three computed anomaly scores is chosen as the final output anomaly score $O_i$ for the data point $X_i$ with
\begin{equation}
\label{eq8}
  O_i = \max\{O_{left}(X_i), O_{right}(X_i), O_{auto}(X_i)\}.
\end{equation}

Algorithm \ref{alg:one} contains the ECOD algorithm's pseudocode, which is utilized in phase 1 of the proposed method. Also, $O_i^{(j)}$ is the dimensional anomaly score for dimension j of $X_i$. Since the log function is monotonic, $O_i^{(j)}$ represents the degree of outlierliness and the anomaly contribution by $X_i^{(j)}$. This representation creates model interpretability for the ECOD model in phase 1 of the method. 

\RestyleAlgo{ruled}
\SetKwComment{Comment}{/* }{ */}
\begin{algorithm}
\caption{Phase 1: Noise Reduction using ECOD.}\label{alg:one}
\textbf{Input:} Training dataset $\{X_i^{(J)}|i=1,2,3...,n\}$ where  $X_i^{(J)}$ refers to the $j-th$ feature (dimension) of the $i-th$ data point\\
\textbf{Output:} Set of anomaly scores $O_i \in R^n$\\

\For {\text{each dimension j in 1,...,d}}{
 Estimate left and right tail ECDFs (using equation \ref{eq2} and \ref{eq3}) \\
 Compute the sample skewness coefficient \{$\Tilde{\mu_3}$\} for the j-th feature's distribution using \eqref{eq4}
}

\For {\text{each sample i in 1,...,n}} {
    Aggregate tail probabilities of $X_i$ to obtain anomaly score $O_i$:\\
    $O_{left}(X_i) = -\sum_{j = 1}^{d} \log(\hat{F}_{left}^{(j)}(X_i^{(j)}))$ \\
    $O_{right}(X_i) = -\sum_{j = 1}^{d} \log(\hat{F}_{right}^{(j)}(X_i^{(j)}))$ \\
    $O_{auto}(X_i) = \sum_{j = 1}^{d}[ \mathbbm{1} \{\Tilde{\mu_3} <0\} \log(\hat{F}_{left}^{(j)}(X_i^{(j)})) +  \mathbbm{1} \{\Tilde{\mu_3} \geq 0\} \log(\hat{F}_{right}^{(j)}(X_i^{(j)}))].$ \\
    Set the final anomaly score for point $X_i$ as: \\
    $O_i = \max\{O_{left}(X_i), O_{right}(X_i), O_{auto}(X_i)\}$. \\
}
\Return{Anomaly scores, $O = (O_1,...,O_n)$}
\end{algorithm}

\subsection{Phase 2: Dual COPOD Model}
\begin{figure*}[h]
\centerline{\includegraphics[width=12cm]{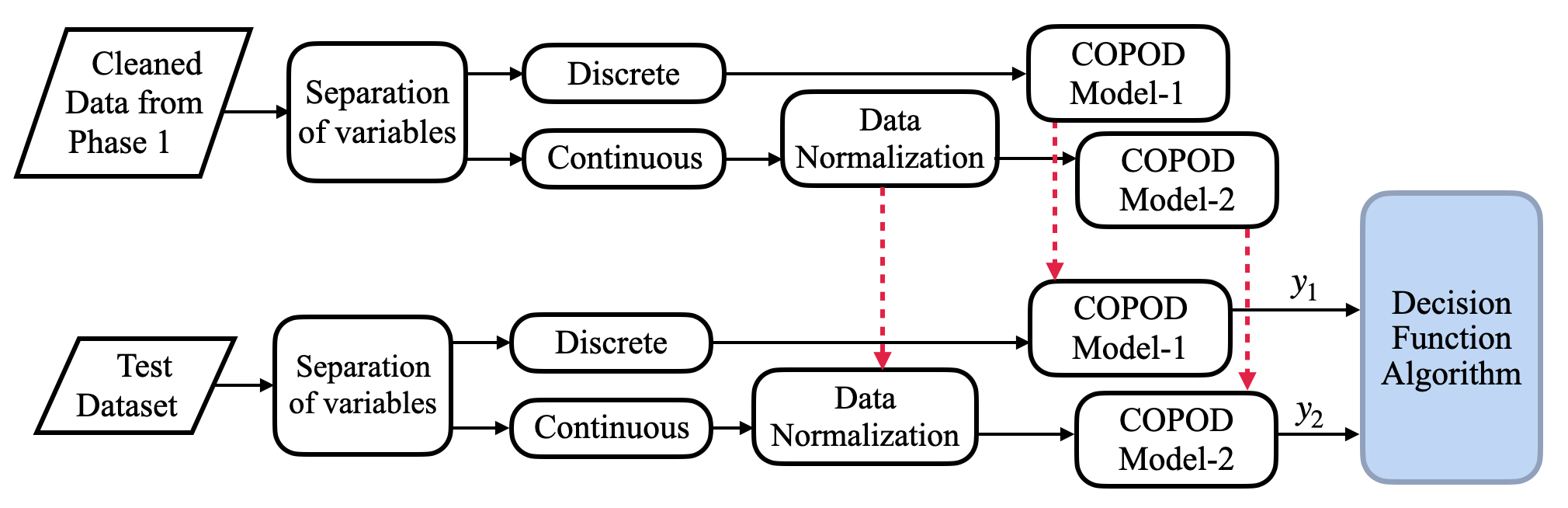}}
\caption{Phase 2 Architecture of the Proposed Method.}
\label{Figure5}
\end{figure*}

Phase 2 of the TPD COPOD consists of two parallel COPOD models. This stage receives the input data from phase 1, which is separated into discrete and continuous data points. One COPOD model receives the discrete input, whereas the other receives the continuous input. Phase 2 of the TPD COPOD model is responsible for the actual anomaly detection after training. The details of the phase 2 architecture are discussed in this section. Figure \ref{Figure5} shows phase 2 architecture of the proposed method. 

Copulas are mathematical operations that allow us to decouple marginal distributions from a given multivariate distribution's dependence structure.  Formally, a d-variate copula $C:[0,1]^{d} \in [0, 1]$, is the CDF of a random vector $(U^{(1)}, U^{(2)},...,U^{(d)})$ with uniform marginals given by
\begin{equation}
\label{eq9}
  C_U(u) = P(U^{(1)} \leq u^{(1)},...,U^{(d)} \leq u^{(d)})
\end{equation}
where $P(U^{(j)} \leq u^{(j)}) = u^{(j)}$ for $j \in 1,...,d$ and $u^{(j)} \in [0, 1]$. By using inverse sampling, uniform distributions can be transformed into any desired distributions, such as
\begin{equation}
\label{eq10}
  X_j = F_j^{-1}(U_j) \sim F_j
\end{equation}

It has been shown in \cite{sklar1959fonctions} that for any random variables $X_1,\cdots,X_d$ with joint distribution function $F(x_1,\cdots,x_d)$ and marginal distributions $F_1,\cdots,F_d$, there exists a copula such that
\begin{equation}
\label{eq11}
  F(x) = C(F_1(x_1),\cdots,F_d(x_d))
\end{equation}
With copula, a joint distribution of $(X_1,\cdots,X_d)$ can be described using only their marginals. This provides flexibility when modeling high-dimensional ICS datasets because the dataset's dimensions can be modeled separately, and a linkage of marginal distributions together to form the joint distribution is gauranteed. Furthermore \cite{sklar1959fonctions} shows that if $F$ has univariate marginal distributions $F_1,\cdots,F_d$, then there exists a copula $C(\cdot)$ such that \eqref{eq11} holds. Additionally, if the marginal distributions are continuous, then $C(\cdot)$ can be uniquely determined. The expression for the copula equation in terms of the joint CDF and inverse CDFs is derived by replacing \eqref{eq9} with the inverse \eqref{eq10} to yield
\begin{equation}
\label{eq12}
  C(u) = F_{x}(F_{x_d}^{-1}(u_1),\cdots,F_{x_d}^{-1}(u_1)).
\end{equation}
Together, \eqref{eq11} and \eqref{eq12} are referred to as Sklar's Theorem, which ensures the existence of a copula for any multivariate CDF with continuous marginals and offers a closed form equation for constructing the copula.

The dual COPOD algorithm of phase 2 uses a nonparametric approach based on fitting ECDFs called empirical copula \cite{li2020copod}. Let $X_{i}^{(j)}$ represent the i-th observation of the j-th feature/dimension for the given d-dimensional ICS training dataset $X$ with $n$ observations. The empirical CDF is
\begin{equation}
\label{eq13}
  \hat{F}(x) = P((-\infty, x]) = \frac{1}{n}\sum_{i = 1}^{n}[ \mathbbm{1}\{X_i \leq x_i\}.
\end{equation}
It is possible to determine  $\hat{U_i}$ by using the empirical copula observations as the inverse of \eqref{eq10}, such that
\begin{equation}
\label{eq14}
  (\hat{U}_{i}^{(1)}, \cdots, \hat{U}_{i}^{d}) = (\hat{F}^{(1)}(X_{i}^{(1)}), \cdots, \hat{F}^{(d)}(X_{i}^{(d)})).
\end{equation}
Finally, by substituting the empirical copula observations of \eqref{eq14} into the first equality of \eqref{eq12}, the empirical copulas are
\begin{equation}
\label{eq15}
  \hat{C}(u^{(1)}, \cdots, u^{(d)}) = \frac{1}{n}\sum_{i = 1}^{n} \mathbbm{1}\{\Tilde{U}_i^{(1)} \leq u^{(1)}, \cdots , \Tilde{U}_i^{(d)} \leq u^{(d)} \}.
\end{equation}
An empirical copula, $\hat{C}(u)$ with multivariate CDFs supported on $n$ points in the grid $\{1/n, 2/n,\cdots, 1 \}^d$, has discrete uniform marginals on $\{1/n, 2/n,\cdots, 1 \}$, and asymptotically converges to $C(u)$ as a result of the central limit theorem \cite{nelsen2007introduction}.

The COPOD algorithm adopts a three-stage process in detecting ICS anomalies. First, the COPOD algorithm computes the ECDFs based on the input dataset received from phase 1. Secondly, the COPOD algorithm uses the ECDFs to produce the empirical copula function. Finally, the COPOD algorithm uses the empirical copula in \eqref{eq15} to approximate the tail probability described in \eqref{eq11}.

The COPOD algorithm's objective is to determine the probability of detecting a point at least as extreme as each observation in the input dataset, $x_i$. That is, assume that $x_i$ is distributed according to d-distribution function $F_X(X_i)$, the COPOD algorithm needs to calculate the tail probabilities, $F_X(x_i) = P(X \leq x_i)$ and $1-F(x_i) = P(X\geq x_i)$. If $x_i$ is an anomaly, the probability of observing a point at least as extreme as $x_i$ should be small. Therefore, if either $F_X(x_i)$ or $1-F(x_i)$ is extremely small, point $x_i$ occurs infrequently and is likely to be an anomaly. In the COPOD algorithm, $F_X(x_i)$ is known as the left tail probability of $x_i$, and  $1-F(x_i)$ is known as the right tail probability of $x_i$. Therefore, an anomaly is considered as an observation that has a small tail probability if either of the two quantities ($F_X(x_i)$ and $1-F(x_i)$) are small.

High-dimensional data spaces have challenges not found in low-dimensional spaces. These challenges are known as the curse of dimensionality, and are common in anomaly detection domain \cite{li2020copod}. In order to prevent diminishing tail probabilities and to exploit the monotonicity property of the $\log()$ function, the COPOD algorithm uses the sum of the negative log probabilities similar to the ECOD algorithm of phase 1 in \eqref{eq5} and \eqref{eq6}. 

\subsection{The Dual COPOD Approach to Anomaly Detection}
The cleaned dataset from phase 1 goes to phase 2 for training the dual COPOD models. In phase 2, the input data is separated into discrete and continuous data points. The discrete data points go to the upper COPOD model (COPOD 1 of Figure \ref{Figure5}), whereas the continuous data is input to the lower COPOD model (COPOD 2 of Figure \ref{Figure5}). The reason for the dual architecture model is to allow phase 2 of the architecture to exploit ICS data by examining the use of two latent representations to extract useful information to minimize overfitting and improve the model's anomaly detection capability. 

Each of the COPOD models in the dual architecture in Figure 
\ref{Figure5} requires a d-dimensional input dataset from phase 1; $X = (X_i^{(1)}, X_i^{(2)},\cdots, X_i^{(d)})$, where $i=1,\cdots,n$, and produces an anomaly score vector $O(X) = [X_1, X_2,\cdots, X_n ]$. The anomaly scores are between $(0,\infty)$, and are to be used comparatively. The anomaly score does not indicate the probability of $X_i$ being an anomaly but rather the relative measure of how likely $X_i$ is when compared to other points in the dataset. Larger $O(X_i)$ signifies $X_i$ is more likely anomalous.

Each COPOD model fits d-dimensional left tail CDFs, using \eqref{eq13} and d-dimensional right tail CDFs by replacing $X$ in \eqref{eq13} with $-X$. Also, d-dimensional skewness vector $\Tilde{\mu_3}$ is computed using \eqref{eq4}. Next, the empirical copula observations for each $X_i$ are computed using \eqref{eq14} to obtain the left tail copulas $\hat{U}_{left}^{j}$ and right tail copulas $\hat{U}_{right}^{j}$. Then, the skewness corrected empirical copula observations are calculated as 
\begin{equation*}
  \hat{W}_i^{(j)} =
  \begin{cases}
  \hat{U}_{left}^{j} & \text{if} \quad \Tilde{\mu_3} < 0 \\ \hat{U}_{right}^{j} & \text{otherwise.} \\ 
  \end{cases}
\end{equation*}
Finally, the probability of observing a point at least as extreme as each $x_i$ along each dimension is computed. Similar to \eqref{eq8}, the maximum of the negative log of the probabilities generated by the left tail empirical copula, right tail empirical copula, and skewness corrected empirical copula is selected as the final anomaly score. That is, the smaller the tail probability is, the bigger its negative log, and so a data point is considered an outlier if it has a small left tail probability, a small right tail probability, or a small skewness corrected tail probability.



 The final anomaly scores of the TPD COPOD are generated as a combination of the anomaly scores of the two COPOD models using a window function. The window function defines the time intervals based on which anomaly scores can be divided using a moving average \cite{hyndman2011moving}. Let $O_{c_1}(X_i)$ and $O_{c_2}(X_i)$ be the anomaly scores of COPOD model 1 and COPOD model 2, respectively. For a given data point $X_i$, its decision label is negative (anomalous) if and only if the predictions of $O_{c_1}(X_i)$ and $O_{c_2}(X_i)$ are negative; otherwise, the decision label of $X_i$ is positive. The final decision score/label $O_f(X_i)$ is made by observing the predictions of the dual COPOD models for a time frame of $t_w$ time instants. In this work, a $t_w$ of 30s is used for inference. Algorithm \ref{alg:three} summarizes the decision function of the TPD COPOD method. Combining the TPD COPOD architecture with the algorithm definition of phase 2 results in the complete TPD COPOD architecture as shown in Figure \ref{Figure5}.

\RestyleAlgo{ruled}
\SetKwComment{Comment}{/* }{ */}
\begin{algorithm}
\caption{TPD COPOD Method's Decision Function.}\label{alg:three}
\textbf{Input 1:} Discrete input data $\{X_{i}^{(J)}|i=1,2,3...,n\}$ where  $X_{i}^{(J)}$ refers to the $j-th$ feature (dimension) of the $i-th$ discrete data point\\
\textbf{Input 2:} Continuous input data $\{X_{i}^{(k)}|i=1,2,3...,n\}$ where  $X_{i}^{(k)}$ refers to the $k-th$ feature (dimension) of the $i-th$ continuous data point\\
\textbf{Output:} Set of decision labels $O_f$ \\

\For {\text{each dimension j in 1,...,d}}{
 Compute anomaly scores $O_{c_1}(X_{i}^{(j)})$ for COPOD 1 
}
\For {\text{each dimension k in 1,...,d}}{
 Compute anomaly scores $O_{c_2}(X_{i}^{(k)})$ for COPOD 2 
}

\For {\text{each data record i in 1,...,n}}{
 \eIf{$O_{c_1}(X_i) \parallel O_{c_2}(X_i) == -1 (anomalous)$}
{
     Output $O_i = 1$; \Comment*[r]{Anomalous data point}
}{
    Output $O_i = 0$ \Comment*[r]{Normal data point}
}
}

\For {\text{each data record i in 1,...,n}}{
 Using a time frame, $t_w$ \\
 \eIf{$\frac{1}{t_w}\sum_{i=1}^{t_w}O_i \geq 80\%$}
{
    Final decision label $O_{f_i} = -1$ \Comment*[r]{Anomaly detected}
}{
     Final decision label $O_{f_i} = 1$ \Comment*[r]{Normal operation}
}
}
\Return{Final decision labels $\{O_{f_i}\}$}
\end{algorithm}

\section{Results and Discussions} \label{result_4}
The TPD COPOD method's performance evaluation is based on performance metrics, results from predictions on the test data, and comparison with prior work trained on similar datasets. The proposed method was developed using the Python programming language, and PyOD which is the most comprehensive and scalable Python library for detecting outlying objects in multivariate data \cite{zhao2019pyod}. Evaluation results and performance metrics calculations are performed by using the Scikit-learn library \cite{scikit-learn}. The three different datasets, namely; SWaT \cite{goh2017anomaly}, WADI \cite{adepu2016generalized}, and TLIGHT \cite{aboah2022plc} are used for evaluating the model performance of the proposed method.


\subsection{Results Evaluation on the SWaT Dataset}
\begin{figure}[h]
\centerline{\includegraphics[width=\columnwidth]{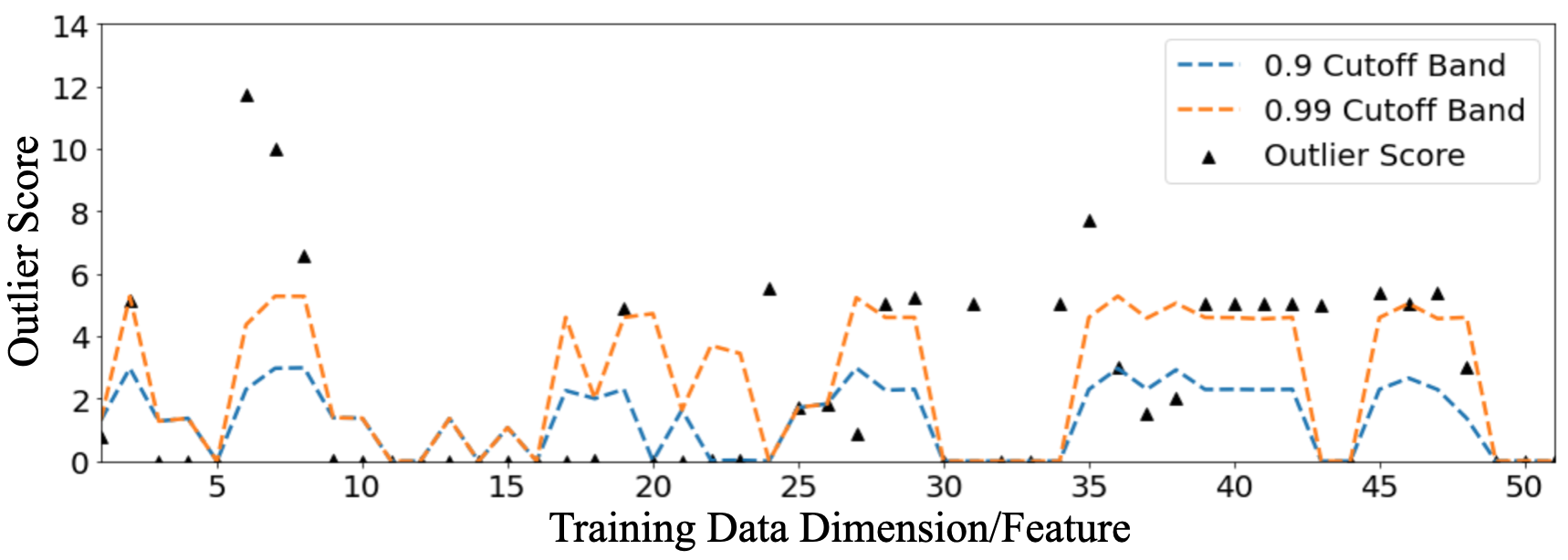}}
\caption{Detected Noise in the   SWaT Training Dataset by Phase 1 shown in Figure \ref{Figure4} of the Proposed Method: Sample 1.}
\label{Figure6}
\end{figure}

This subsection describes the proposed method's performance on the SWaT dataset as compared with previous work evaluated on the same dataset. During training, the SWaT training dataset is passed through the first stage of the TPD COPOD method for any noise and unwanted signals to be removed. Figure \ref{Figure6} shows the first sample of noise detected by the ECOD model of phase 1 TPD COPOD. Figure \ref{Figure6} shows the feature-level outlier scores explaining the reason for detecting the first sample as an outlier. The x-axis indicates the features (sensors and actuators) of the input dataset represented as numerical integers. The blue dashed line represents the 90th percentile band, and the orange dashed line represents the 99th percentile band for the given features. Sample 1 of Figure \ref{Figure6} is flagged as an anomaly because several dimensions (features), such as dimensions 2, 6-8, 19, 24, 28, 29, 31, 34, 35 39-43, and 45-48, have outlier scores that exceed the 99th percentile band. The detected anomalous training sample 1 shown in Figure \ref{Figure6} is selected at random, and the explanations provided about the figure is similar to the rest of the plots, and are not shown in this paper. Figure \ref{Figure6} shows that about half of the dimensions have outlier scores that exceed the 99th percentile band, and this explains why the sample is detected as an anomaly. Overall, the computer used for analysis required 16.70s for phase 1 to clean the training data consisting of 496,800 samples. A total of 49,680 training samples were detected as outliers in phase 1 of TPD COPOD. 

\begin{figure}[h]
\centerline{\includegraphics[width=\columnwidth]{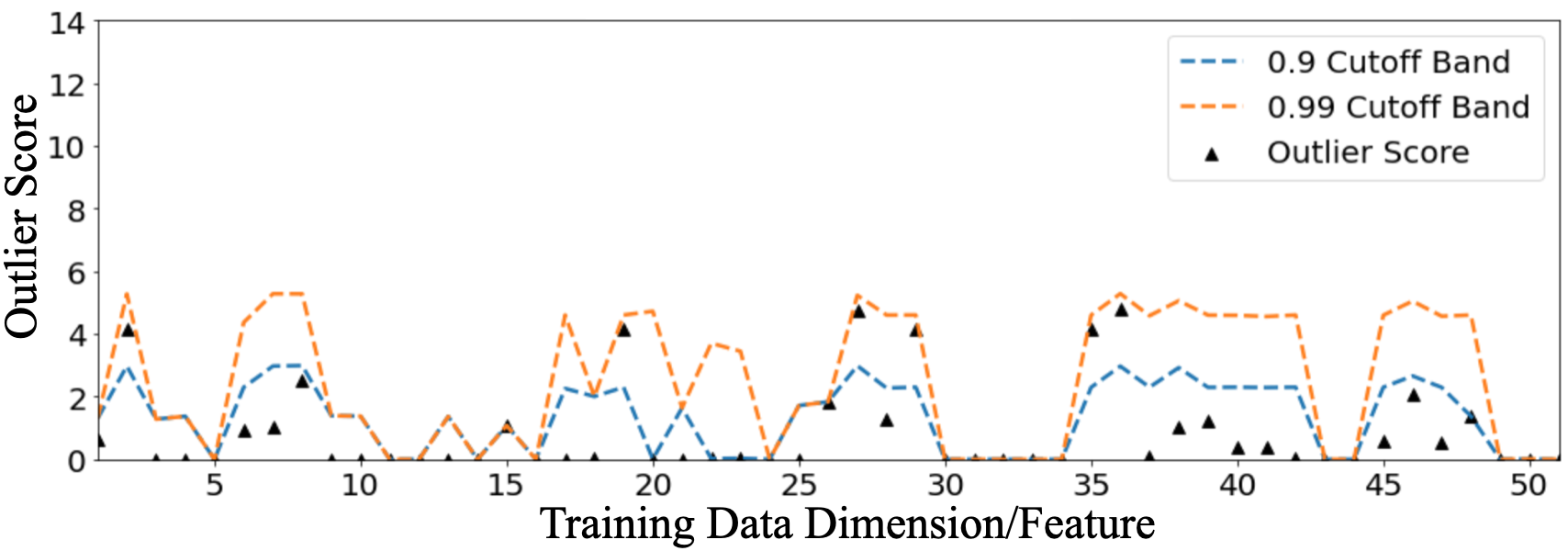}}
\caption{Detected Normal Data Sample in the SWaT Training Dataset by Phase 1 shown in Figure \ref{Figure4} of the Proposed Method: Sample 1.}
\label{swat_normal1}
\end{figure}

Figure \ref{swat_normal1} shows the first normal training sample in the SWaT training dataset detected by the ECOD model of phase 1 TPD COPOD model. The training sample shown in Figure \ref{swat_normal1} is detected as a normal sample because none of the dimensional outlier scores exceeds the 99th percentile band. Similarly, in all cases where the ECOD model of phase 1 TPD COPOD model detected a training sample as normal, none of the dimensional anomaly scores of those samples exceeded the 99th percentile band.

\begin{figure}[h]
\centerline{\includegraphics[width=\columnwidth]{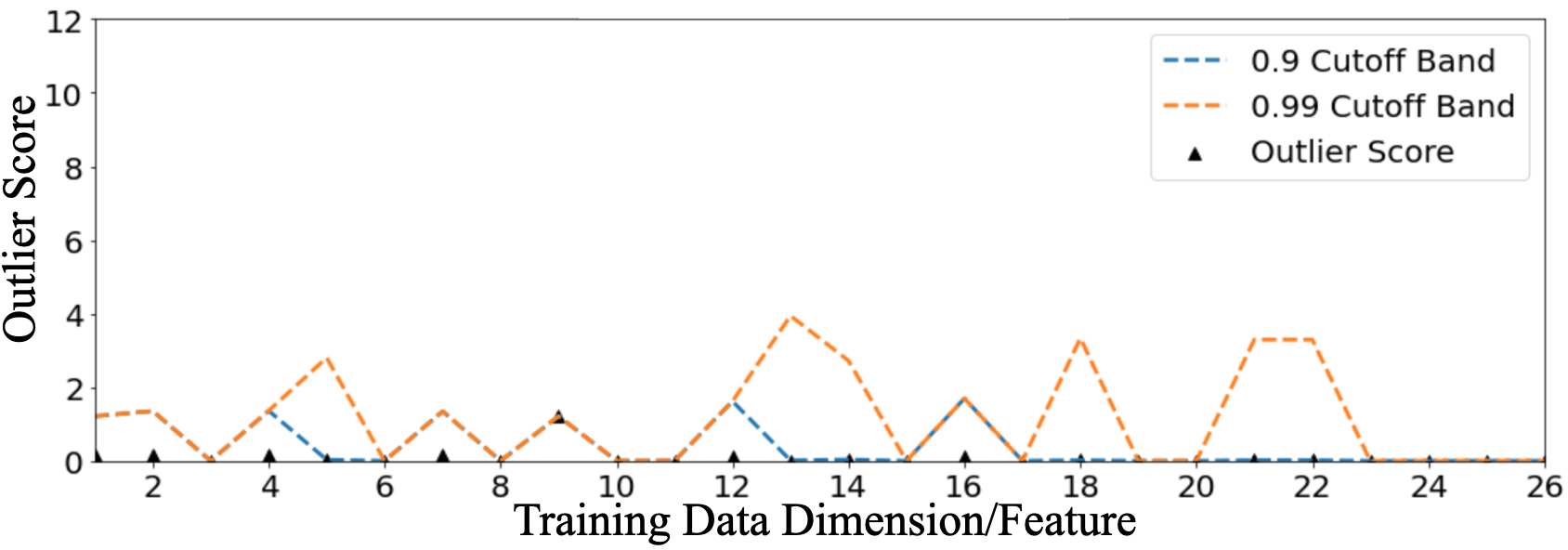}}
\caption{Detected Anomaly in the SWaT Test Dataset by Phase 2 shown in Figure \ref{Figure5} of the Proposed Method: Discrete Sample 1.}
\label{Figure11}
\end{figure}

The clean SWaT dataset consisting of 447,120 samples were separated into discrete and continuous samples for phase 2 of the TPD COPOD. Training required 2.35s for phase 2 models, and predictions required 6.02s over test data consisting of 450,819 samples. After training, only phase 2 of the proposed method is used for performing inference. Figure \ref{Figure11} shows the first discrete sample of a detected anomaly in the test data by phase 2 COPOD 1. Figure \ref{Figure11} shows that because of the discrete nature of the input dataset to phase 2 COPOD 1, the 90th percentile band for most of the dimensions are 0. Dimension 9 produced an outlier score that is equal to the 99th percentile band, and as a result, dimension 9 is a major contributing factor to classifying the first discrete test sample as an anomaly.

\begin{figure}[h]
\centerline{\includegraphics[width=6cm]{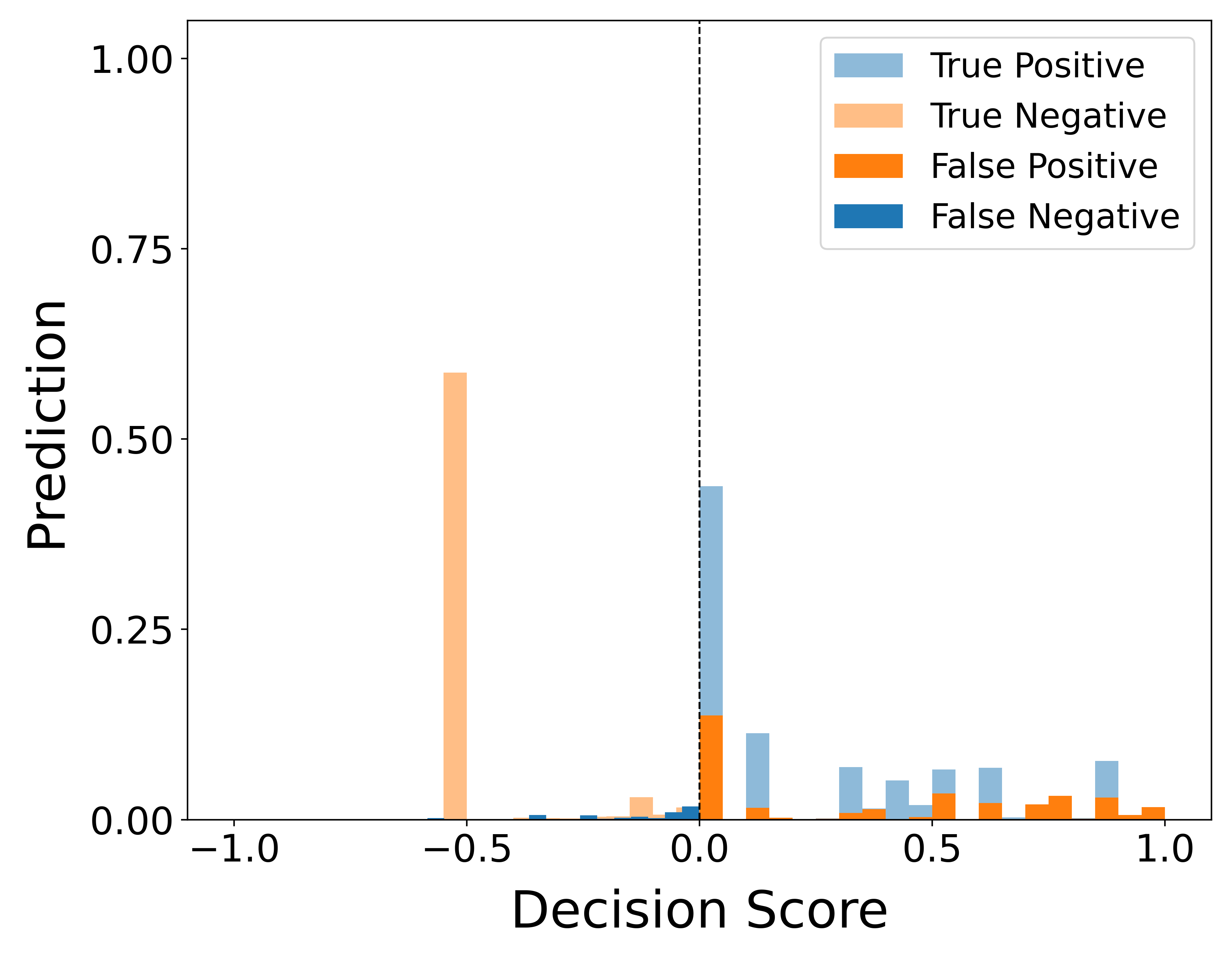}}
\caption{Results of the Normalized TP, TN, FP, and FN Values on the SWaT Test Set by Phase 2 COPOD 1 shown in Figure \ref{Figure5}.}
\label{swat_discrete}
\end{figure}

\begin{figure}[h]
\centerline{\includegraphics[width=6cm]{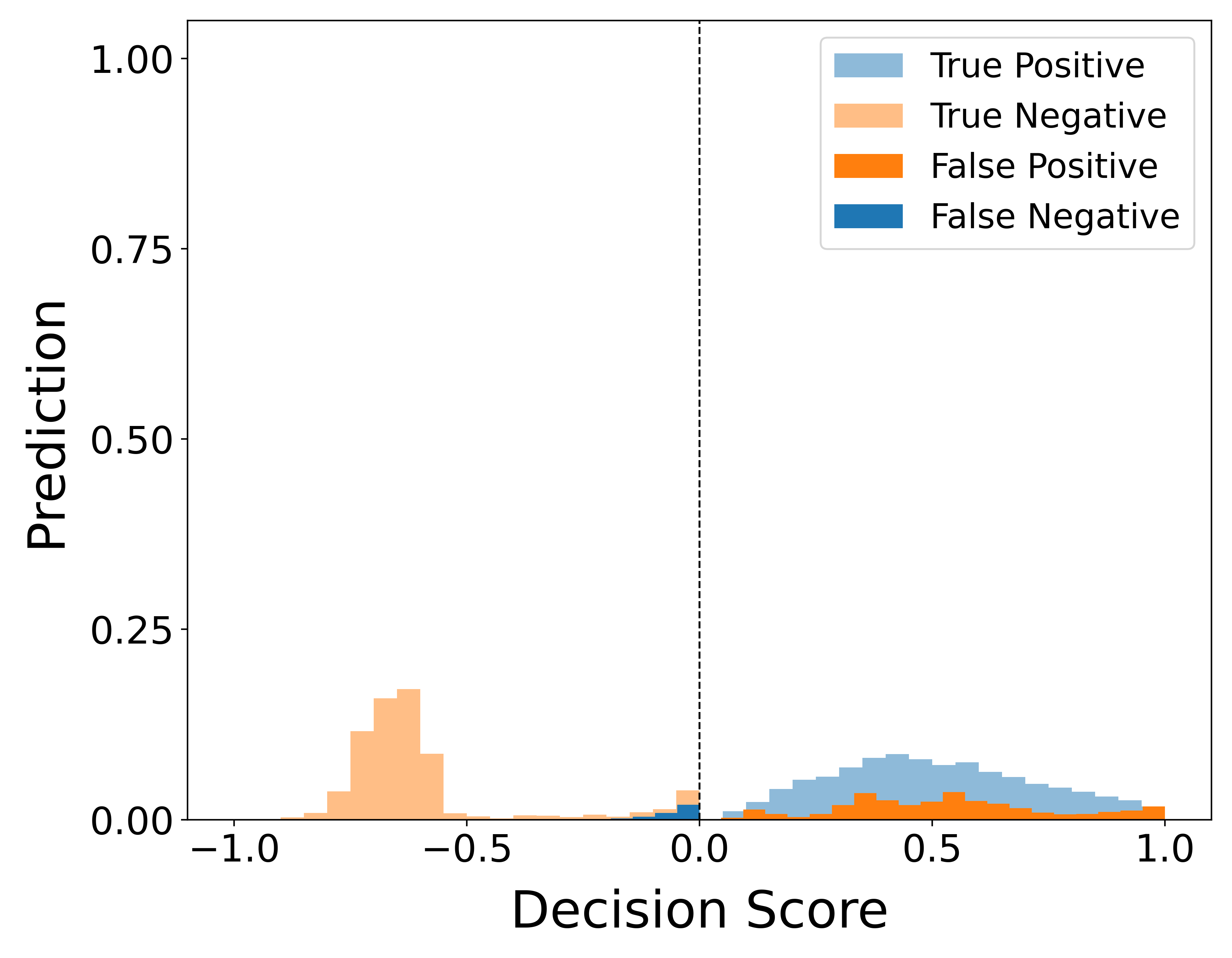}}
\caption{Results of the Normalized TP, TN, FP, and FN Values on the SWaT Test Set by Phase 2 COPOD 2 shown in Figure \ref{Figure5}.}
\label{swat_continuous}
\end{figure}

The results presented in Figure \ref{swat_discrete} use the visualization approach proposed in \cite{aboah2022plc, aboahboateng2022new} to better understand the TPD COPOD model performance. The histogram-based visualization approach normalizes the histogram frequency (y-axis) to a range between $0\%$ and $100\%$, whereas the x-axis represents the model's normalized decision scores indicating the prediction confidence. Figure \ref{swat_discrete} shows the results of the normalized TP, TN, FP, and FN values on the SWaT test set by phase 2 COPOD 1 shown in Figure \ref{Figure5}. Figure \ref{swat_discrete} shows that phase 2 COPOD 1 correctly detected about 60\% of the SWaT anomalies with over 60\% confidence level. Phase 2 COOPD 1 misclassified less than 10\% of the normal instance as anomalies, whereas about 35\% of the anomalies were misclassified as normal data points. Also, phase 2 COPOD 1 correctly predicted about 45\% of the normal data with prediction confidence of less than 10\%. Figure \ref{swat_continuous} shows the results of the normalized TP, TN, FP, and FN values on the SWaT test set by phase 2 COPOD 2 shown in Figure \ref{Figure5}. Figure \ref{swat_continuous} shows that about 80\% of the anomalies were correctly detected by phase 2 COPOD 2 with prediction confidence between 95\% and 50\%. The TP and FP prediction by phase 2 COPOD are normally distributed with average prediction confidence of about 50\%. 

The SWaT dataset has been utilized for anomaly detection by K-Nearest Neighbors (KNN) \cite{li2018anomaly}, Feature Bagging (FB) \cite{li2018anomaly}, Support Vector Machines (SVM) \cite{li2018anomaly}, Auto-encoders (AE) \cite{li2018anomaly} and Dual Isolation Forest (DIF) \cite{elnour2020dual}.
Table \ref{tab2} shows a comparison of evaluation results between previously published methods and the proposed method. Table \ref{tab2} shows that TPD COPOD has an F1-score of $93\%$, precision of $93\%$, and recall of $93\%$. TPD COPOD has the highest F1-score and recall values as compared with previous work. In terms of precision, TPD COPOD performed at par with NN-one class, DIF, and SVM. The high recall and F1-score values of TPD COPOD reflect its robustness and ability to confidently detect SWaT dataset anomalies compared to previous work. When compared to other approaches with similar computational complexities, such as NN, the proposed anomaly detection method significantly outperforms NN in terms of F1-score and recall. The outstanding performance of TPD COPOD across all performance metrics could be attributed to the noise removal from the training data in phase 1 during model training. Phase 2 of the method in Figure \ref{Figure5} trains on clean training data, arguably a true representation of the normal operations of the SWaT testbed.

\begin{table}
    \centering
    \caption{Comparison between Detection Methods Evaluated on the SWaT Dataset}
    \label{tab2}
    \setlength{\tabcolsep}{3pt}
    \begin{tabular}{|p{2.2cm}|p{1.3cm}|p{1cm}|p{1cm}|p{1.8cm}|}
    \hline
     Method & F1-Score & Precision & Recall & Complexity \\
    \hline
    NN \cite{shalyga2018anomaly} &  0.812 & \textbf{0.976}	& 0.696	& Low\\
    RNN \cite{inoue2017anomaly} & 0.802 &	0.982 &	0.678 &	High\\
    SVM \cite{inoue2017anomaly} & 0.796 &	0.925 &	0.699 &	High\\
    TABOR(TB) \cite{lin2018tabor} & 0.823	& 0.862	& 0.788	& Average\\
    ID-CNN \cite{kravchik2018detecting} &0.860 &	0.867 &	0.854 &	High\\
    AE \cite{li2018anomaly} & 0.520 & 0.516	& 0.516	& Average\\
    FB \cite{li2018anomaly}	& 0.360 &	0.358 &	0.358 &	Average\\
    KNN \cite{li2018anomaly} & 0.350	& 0.348	& 0.348	& Average\\
    GAN \cite{li2018anomaly} & 0.510 & 0.406 & 0.677 & High\\
    DIF \cite{elnour2020dual} & 0.882	& 0.935	& 0.835	& Average\\
    NN-one class \cite{boateng2022anomaly} & 0.870 & 0.940 & 0.820 & Average\\
    \textbf{TPD COPOD} & \textbf{0.930} & 0.930 & \textbf{0.930} & \textbf{Very low}\\
    \hline
    \end{tabular}
\end{table}

\subsection{Results Evaluation on the WADI Dataset}
This subsection describes the proposed method's performance on the WADI dataset compared to previous work evaluated on the same dataset. During training, the WADI training dataset is passed through the first stage of the TPD COPOD method for any noise and unwanted signals to be removed. The total number of dimensions (features) consisting of both discrete and continuous values of the training dataset is 119. The clean WADI data consisting of 706,114 samples were separated into discrete and continuous samples for phase 2 of the TPD COPOD. The computer used for analysis required 13.80s to train the phase 2 models and 19.64s to make predictions on the test data consisting of 172800 samples.

The WADI dataset has been utilized for anomaly detection by SVM \cite{li2018anomaly} KNN \cite{li2018anomaly}, AE \cite{li2018anomaly}, FB \cite{li2018anomaly}, EGAN \cite{li2018anomaly}, GAN \cite{li2018anomaly}, Deep Autoencoding Gaussian Model (DAGMM) \cite{zong2018deep}, Long-short Term Memory Variational Autoencoder (LSTM-VAE) \cite{park2018multimodal} and DIF \cite{elnour2020dual}, and Graph Deviation Network (GDN) \cite{deng2021graph}. Table \ref{tab3} shows a comparison of evaluation results between previous work and the proposed method. Table \ref{tab3} shows that TPD COPOD has superior performance to all previous work in terms of F1-score and recall. TPD COPOD achieved an F1-score of $92\%$, precision of $82\%$, and recall of $86\%$. The high recall and F1-score values of TPD COPOD reflect its robustness and ability to confidently detect WADI dataset anomalies as compared to previous work. Although LSTM-VAE and GDN achieved higher precision values as compared to the proposed method in this study, LSTM-VAE and GDN models have significantly low recall values of $27\%$ and $40.2\%$, respectively. The low recall values of LSTM-VAE and GDN resulted in poor F1-scores of $25\%$ and $57\%$, respectively. The outstanding performance of TPD COPOD across all performance metrics could be attributed to the noise removal from the training data in phase 1 during model training.  Phase 2 of the method in Figure \ref{Figure5} trains on clean training data, arguably a true representation of the normal operation of the WADI testbed.

\begin{table}
    \centering
    \caption{Comparison between Detection Methods Evaluated on the WADI Dataset}
    \label{tab3}
    \begin{tabular}{|p{2cm}|p{1.2cm}|p{1.05cm}|p{1cm}|p{1.5cm}|}
    \hline  
     Method & F1-Score & Precision & Recall & Complexity \\
    \hline
    PCA \cite{shalyga2018anomaly} &  0.250 & 0.504	& 0.166 & Average	\\
    SVM \cite{inoue2017anomaly} & 0.510 &	0.512 &	0.512 & High\\
    AE \cite{li2018anomaly} & 0.520 & 0.520	& 0.520 & Average \\
    FB \cite{li2018anomaly}	& 0.340 &	0.336 &	0.336 & Average\\
    KNN \cite{li2018anomaly} & 0.300	& 0.299	& 0.299	& Average\\
    EGAN \cite{li2018anomaly} & 0.340 & 0.345 & 0.345 & High\\
    GAN \cite{li2018anomaly} & 0.620 & 0.538 & 0.749 & High\\
    DIF \cite{elnour2020dual} & 0.656	& 0.765	& 0.574	& Average\\
    DAGMM \cite{zong2018deep} & 0.360  &  0.544& 0.270 & Average\\
    LSTM-VAE \cite{park2018multimodal} & 0.250 &  0.878 & 0.145 & High\\
    GDN \cite{deng2021graph} & 0.570 &  \textbf{0.975} &  0.402 & High\\
    \textbf{TPD COPOD} & \textbf{0.920} & 0.820 & \textbf{0.860} & \textbf{Very low} \\
    \hline
    \end{tabular}
\end{table}

\subsection{Results Evaluation on the TLIGHT Dataset}
This subsection describes the proposed method's performance on the TLIGHT dataset as compared with previous work evaluated on the same dataset. During training, the TLIGHT training dataset is passed through the first stage of the TPD COPOD method for any noise and unwanted signals to be removed. The total number of dimensions (features) consisting of both discrete and continuous values of the training dataset is 36. There are five different test sets of TLIGHT dataset. The computer used for analysis required 0.67s for phase 1 to clean the training data consisting of 41,580 samples. A total of 4,144 training samples were detected as outliers in phase 1.

Table \ref{tab4} shows the comparison between TPD COPOD and previous methods evaluated on TLIGHT dataset. Test set 1 of the TLIGHT dataset consisted of 5,000 samples, the computer required 0.328s to make predictions. Table \ref{tab4} shows that TPD COPOD achieves superior precision, recall, and F1-score of $96\%$, $96\%$, and $95\%$ respectively. OCSVM \cite{aboah2022plc}, OCNN \cite{aboah2022plc} and IF \cite{aboah2022plc} have similar performance in terms of precision, recall, and F1-score. The test set 2 of the TLIGHT dataset consisted of 7,000 samples. The computer used for analysis required 0.0.437s to make predictions on the test set 2. TPD COPOD achieves superior performance by correctly predicting all the data points in test set 2 by achieving $100\%$ precision, recall, and F1-score. OCSVM, OCNN and IF have similar performance in terms of precision, recall, and F1-score. IF achieves high precision, recall, and F1-score values of $98\%$, $97\%$, and $97\%$ respectively, on test set 2. However, OCNN and OCSVM achieved similar performance on test set 2. The test set 3 of the TLIGHT dataset consisted of 13,130 samples. The computer used for analysis required 0.324s to make predictions on the test set 3. TPD COPOD achieves superior precsion, recall, and F1-score of $91\%$, $90\%$, and $88\%$ respectively, on test set 3. OCSVM, OCNN and IF achieve similar performance in terms of precision, recall, and F1-score on test set 3. 

The test set 4 of the TLIGHT dataset consisted of 15,000 samples. The computer used for analysis required 0.477s to make predictions on the test set 4. TPD COPOD achieved relatively low performance on test set 4 as compared to its performance on test set 1 and 2 by having precision, recall, and F1-score values of $88\%$, $85\%$, and $83\%$ respectively. OCNN, IF, and TPD COPOD achieve similar performance whereas OCSVM has the worst performance by having precision, recall, and F1-score values of $81\%$, $81\%$, and $71\%$ respectively. TPD COPOD has low performance on test set 4 because of the large proportion of anomalies in test set 4 consisting of timing bits anomalies which are hard to detect \cite{aboah2022plc}. The test set 5 of the TLIGHT dataset consisted of 18,269 samples. The computer used for analysis required 0.517 to make predictions on the test set 5. TPD COPOD achieved its lowest performance on test set 5 by having precision, recall, and F1-score values of $83\%$, $75\%$, and $73\%$ respectively. OCNN and OCSVM achieve similar performance whereas IF has the best performance by having precision, recall, and F1-score values of $82\%$, $78\%$, and $77\%$ respectively. Again, TPD COPOD has low performance on test set 5 because of the large proportion of anomalies in test set 5 consisting of timing bits anomalies. Therefore, the TPD COPOD appears to be ineffective at detecting TLIGHT system errors consisting of system timing bits. 

\begin{table}
\begin{center}
\caption{Comparison between Detection Methods Evaluated on TLIGHT Dataset}
\label{tab4}
\begin{tabular}{| p{0.7cm} | p{1.7cm} | p{0.9cm} | p{0.8cm} | p{0.7cm} | p{1.3cm} |}
\hline
Dataset & Method & Precision & Recall & F1-Score & Complexity\\
\hline
  Test & OCSVM &0.89  &0.90  & 0.98 & High\\
  Set & OCNN &0.90  &0.91  & 0.90 & Average\\
  1 & IF &0.90  &0.91  & 0.91 & Average\\
   & \textbf{TPD COPOD}   & \textbf{0.96}  & \textbf{0.96}  & \textbf{0.95} & \textbf{Very low}\\
   \hline
     Test &OCSVM &0.81  &0.87  & 0.89 & High\\
   Set & OCNN &0.81  &0.88  & 0.84 & Average\\
   2 & IF &0.98  &0.97  & 0.97 & Average\\
   & \textbf{TPD COPOD}   & \textbf{1.00}  & \textbf{1.00}  & \textbf{1.00} & \textbf{Very low}\\
   \hline
    Test & OCSVM &0.85  &0.86  & 0.84 & High\\
   Set & OCNN &0.87  &0.88  & 0.86 & Average\\
   3 & IF &0.87  &0.87  & 0.87 & Average\\
   & \textbf{TPD COPOD}   & \textbf{0.91}  & \textbf{0.90}  & \textbf{0.88} & \textbf{Very low}\\
   \hline
     Test & OCSVM &0.81  &0.81  & 0.71 & High\\
   Set & OCNN &0.87  & \textbf{0.88}  & \textbf{0.86} & Average\\
   4 & IF &0.86  &0.86  & 0.85 & Average\\
   & \textbf{TPD COPOD}   & \textbf{0.88}  &0.85  & 0.83 & \textbf{Very low}\\
   \hline
     Test & OCSVM &0.78  &0.70  & 0.68 & High\\
   Set & OCNN &0.79  &0.70  & 0.68 & Average\\
   5 & IF &0.82  & \textbf{0.78}  & \textbf{0.77} & Average\\
   & \textbf{TPD COPOD}   & \textbf{0.83}  &0.75  & 0.73  & \textbf{Very low}\\
\hline
\end{tabular}
\end{center}
\end{table}

\section{Conclusion} \label{conclusion_4}
This work proposes the first known two-phase dual (TPD) COPOD anomaly detection method, which is an unsupervised anomaly detection technique consisting of two sequential stages and a dual (parallel) modeling stage. Phase 1 of the method makes use of ECOD to remove any obvious noise or outlier data records in a given training dataset. Phase 2 of the method consists of a dual COPOD architecture that utilizes the output data of phase 1 to develop two COPOD models. The algorithms implemented in the proposed method are parameter-free and based on empirical distribution functions. The deterministic nature of all stages of the method results in the mitigation of the challenges associated with hyperparameter selection in unsupervised anomaly detection. Furthermore, the proposed method is highly interpretable and quantifies each feature's contribution toward an ICS anomaly. The proposed anomaly detection method is computationally and memory efficient, scalable, and suitable for low- and high-dimensional ICS datasets. The proposed method is trained, evaluated, and compared with previous work using three open-source ICS datasets. The proposed method outperformed previous work in terms of F1-score and recall on the SWaT, WADI and TLIGHT datasets. The robust performance of the TPD COPOD method coupled with its speed of anomaly detection makes the TPD COPOD capable of real-time ICS anomaly detection. Future work should focus on finding a means by which the ECOD algorithm of phase 1 may be extended to multimodal training dataset.

\bibliographystyle{IEEEtran}
\bibliography{bare_jrnl_new_sample4}

\end{document}